\documentclass[10pt,twocolumn,letterpaper]{article}

\usepackage{cvpr}              
\definecolor{cvprblue}{rgb}{0.21,0.49,0.74}
\usepackage[pagebackref,breaklinks,colorlinks,allcolors=cvprblue]{hyperref}

\usepackage{amsmath,amssymb,amsfonts}
\usepackage{algorithm}
\usepackage[noend]{algpseudocode}
\algrenewtext{For}[1]{\textbf{For} #1}
\algrenewtext{EndFor}{\textbf{end for}}
\usepackage{amsmath}
\usepackage{algorithm}
\usepackage{algpseudocode}

\usepackage{graphicx}
\usepackage{amsmath}
\usepackage{amssymb}
\usepackage{booktabs}
\usepackage{wrapfig}
\usepackage{multirow}  %
\usepackage{graphicx}  %
\usepackage{diagbox}   %
\usepackage{stfloats}  %

\usepackage{tabularray} %

\usepackage{colortbl}
\usepackage{xcolor}
\definecolor{RowColor}{rgb}{0.93, 0.93, 1}

\usepackage{float}
\usepackage{balance}

\def\paperID{} 
\def\confName{CVPR}
\def\confYear{2026}

\title{Den-TP: A \underline{Den}sity-Balanced Data Curation and Evaluation Framework for \underline{T}rajectory \underline{P}rediction}

\author{Ruining Yang \quad Yi Xu \quad Yun Fu \quad Lili Su\\
Northeastern University, USA \\
{\tt\small \{yang.ruini, xu.yi, y.fu, l.su\}@northeastern.edu}
}

\begin{document}
\maketitle
\begin{abstract}
Trajectory prediction in autonomous driving has traditionally been studied from a model-centric perspective.
However, existing datasets exhibit a strong long-tail distribution in scenario density, where common low-density cases dominate and safety-critical high-density cases are severely underrepresented.
This imbalance limits model robustness and hides failure modes when standard evaluations average errors across all scenarios. 
We revisit trajectory prediction from a data-centric perspective and present Den-TP, a framework for density-aware dataset curation and evaluation. 
Den-TP first partitions data into density-conditioned regions using agent count as a dataset-agnostic proxy for interaction complexity. 
It then applies a gradient-based submodular selection objective to choose representative samples within each region while explicitly rebalancing across densities.
The resulting subset reduces the dataset size by 50\% yet preserves overall performance and significantly improves robustness in high-density scenarios. 
We further introduce density-conditioned evaluation protocols that reveal long-tail failure modes overlooked by conventional metrics. 
Experiments on Argoverse 1 and 2 with state-of-the-art models show that robust trajectory prediction depends not only on data scale, but also on balancing scenario density.

\end{abstract}
    
\section{Introduction}
\label{sec:intro}

Trajectory prediction plays a key role in autonomous driving. With rapid developments in deep learning, various methods~\citep{zhou2022hivt,feng2023macformer,chai2019multipath,gu2021densetnt,zhang2021map,zhao2021tnt,ngiam2021scene,xu2022adaptive,xu2023uncovering,xu2025vlm} have been proposed and achieved strong performance, but their performance across different types of driving scenarios remains poorly understood. 
This gap is partly driven by the structure of modern trajectory datasets which, despite their scale, exhibit a pronounced long-tail imbalance in scenario density.

\begin{figure}[t]
\centering
\includegraphics[width=\linewidth]{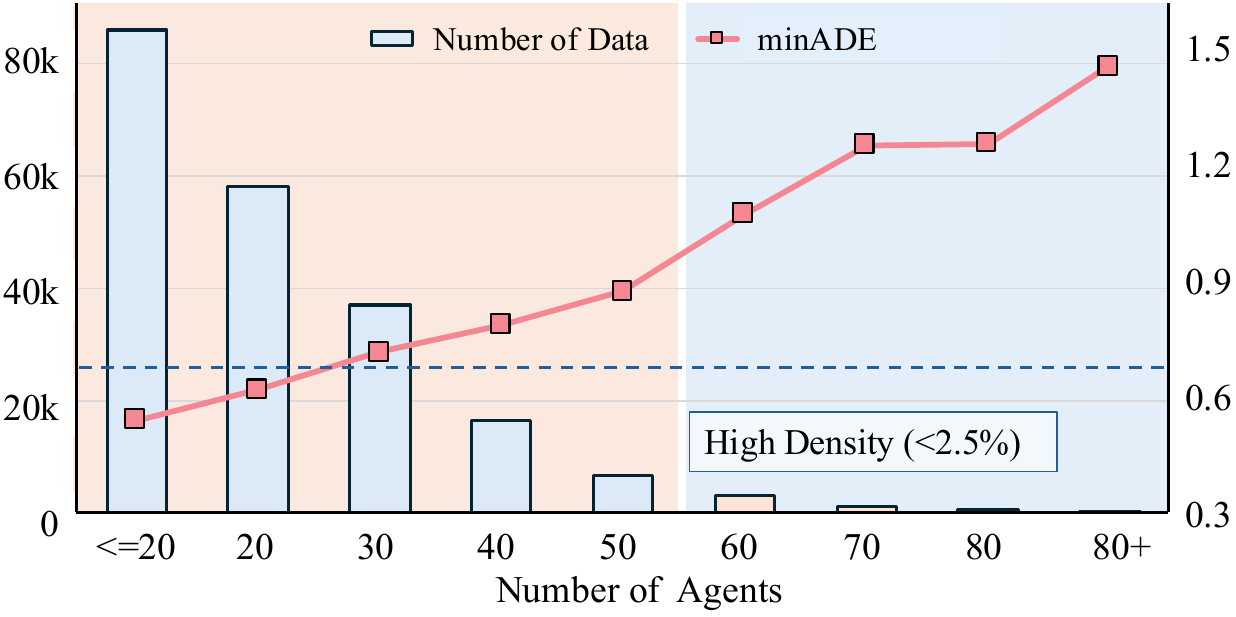}
\caption{\footnotesize Visualization of scenario density distribution and density-conditioned HiVT-64 prediction performance on Argoverse 1.  
Bars (left axis) show that the dataset is heavily skewed toward low-density scenarios, while high-density scenarios (\textgreater 60 agents) account for fewer than 2.5\% of samples.  
The minADE curve (right axis) reveals a monotonic increase in prediction error with density.  
The dashed horizontal line represents the minADE over the full dataset and illustrates how aggregate metrics are dominated by abundant low-density scenarios, masking errors in rare but safety-critical high-density scenarios.}
\label{fig:fig1}
\vspace{1em}
\end{figure}

Across Argoverse 1~\citep{chang2019argoverse}, Argoverse 2~\citep{wilson2023argoverse}, and other driving datasets~\citep{caesar2020nuscenes,zhan2019interaction,sun2020scalability}, we observe a consistent pattern:
low-density scenarios constitute the vast majority of the data, whereas complex high-density interactions are severely underrepresented. Yet such high-density cases are among the most safety-critical for autonomous driving, since they involve complex multi-agent interactions where small prediction errors may directly compromise driving safety. Similar long-tail phenomena have been observed in prior work~\citep{makansi2021exposing,wang2023fend,pourkeshavarz2023learn,chen2024criteria}. However, high-density scenarios contribute little to the overall training signal and are diluted under standard evaluation protocols that average errors over the full dataset.
Standard trajectory prediction pipelines typically assume that all scenarios contribute equally and therefore rely on uniform sampling during training.
This assumption overlooks a central challenge in driving datasets: model performance varies substantially across density regions, and rare high-density scenarios receive insufficient supervision due to their limited representation.
Moreover, current evaluation practices seldom expose this imbalance, as density-conditioned metrics are rarely reported and aggregated performance often masks degradation in safety-critical regions.
To address this issue, we adopt a data-centric perspective.
We treat scenario density as a conditioning variable that shapes the distribution of interaction patterns.
Although agent count does not fully capture the complexity of a scenario, it provides a dataset-agnostic proxy that enables consistent cross-dataset analysis.
This perspective allows us to expose the bias induced by density imbalance and to reevaluate prediction performance under more informative, density-conditioned protocols.

Building on this perspective, we introduce Den-TP, a framework for density-aware dataset curation and evaluation for trajectory prediction. 
Specifically, Den-TP consists of two main stages: extraction and selection. 
In the \textbf{extraction stage}, a lightweight pretraining step is used to obtain stable gradient estimates, and the dataset is partitioned by the number of agents into density ranges, yielding an interpretable structure over the dataset.
In the \textbf{selection stage}, within each density region, we apply submodular selection based on gradient-based influence scores to retain representative samples. Across regions, we employ biased sampling to explicitly upweight rare but critical high-density scenarios, preventing dominance by the majority of low-density scenarios. 
Experiments demonstrate that Den-TP successfully constructs a compact and more balanced subset that retains only half of the original data, yet maintains overall performance and yields clear gains in high-density scenarios. 
Incorporating scenario density into the data selection process helps alleviate the systematic undertraining of rare, interaction-heavy cases. 
Moreover, our framework introduces density-conditioned evaluation protocols that expose long-tail failure modes obscured by conventional aggregate metrics. 
These findings show that robust trajectory prediction depends not only on dataset scale but also on balanced coverage across density regions, underscoring the importance of data-centric curation and evaluation. The main contributions are summarized as follows: 
\begin{itemize}
\item We reveal that modern trajectory prediction datasets are heavily skewed across density regions, which biases models toward majority cases and limits performance in safety-critical rare scenarios; this imbalance induces large performance gaps that are largely hidden under standard aggregate evaluation.

\item We introduce Den-TP, a density-aware curation framework that partitions data by scenario density and selects representative samples using submodular objectives and gradient-based influence scores, while biased sampling preserves rare safety-critical cases and density-conditioned evaluation protocols expose their associated failure modes.

\item By selecting only 50\% of the data, Den-TP constructs a compact and density-balanced subset that maintains overall performance, substantially reduces computational cost, improves performance in high-density interaction scenarios, and transfers robustly across diverse model architectures.
\end{itemize}

\begin{figure}[t]
\centering
\includegraphics[width=\linewidth]{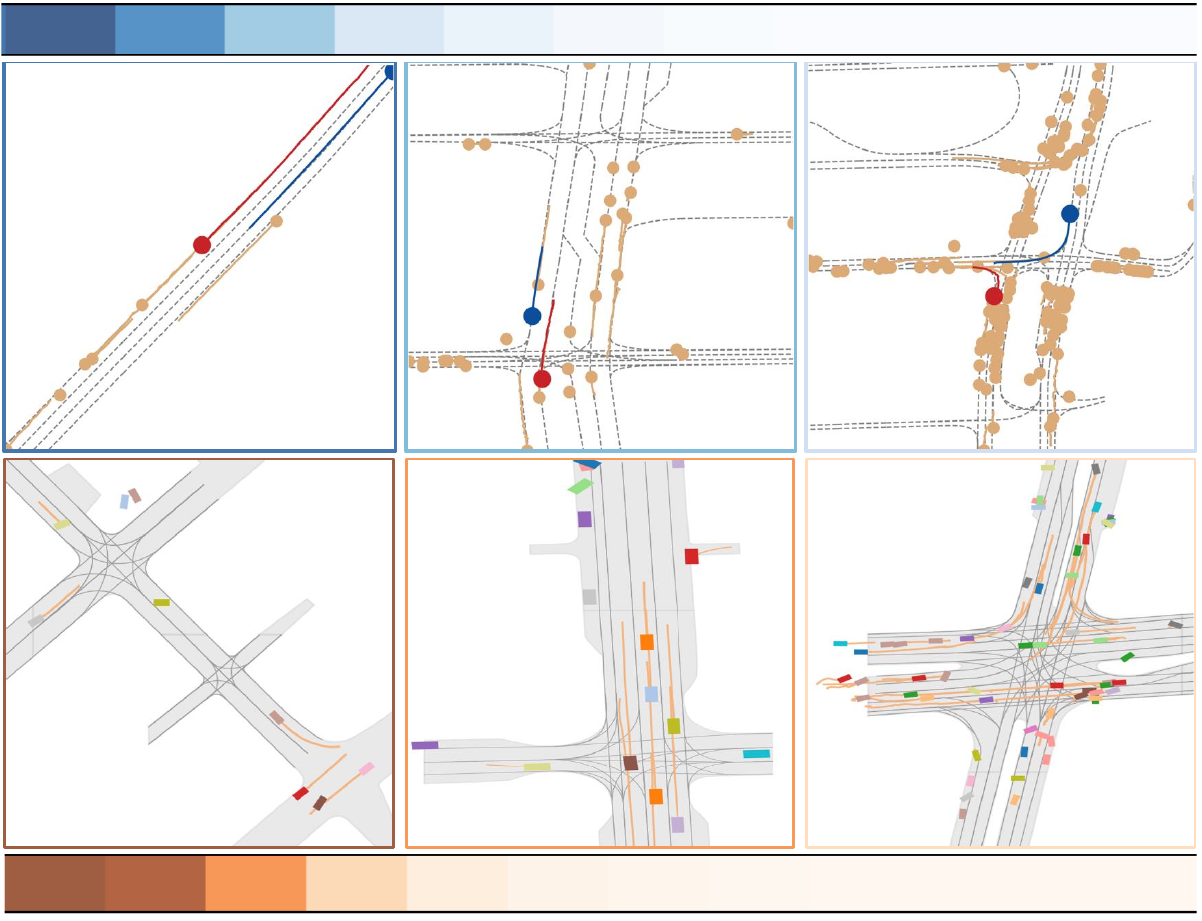
}
\caption{\footnotesize Visualization of scenarios with different densities in Argoverse 1 (top) and Argoverse 2 (bottom). The heat map indicate sample frequency, where darker colors represent density ranges that appear frequently in the dataset, and lighter colors correspond to rare density ranges. From left to right, the illustrated scenarios progress from low-density to high-density settings. Both datasets exhibit the same long-tail pattern that most scenarios fall into low-density regions, while high-density are less common.}
\label{fig:fig2}
\vspace{1em}
\end{figure}

\section{Related Work}
\label{sec:related_work}

\noindent\textbf{Scenario Imbalance in Trajectory Data.}
Trajectory prediction has made rapid progress through Transformer-based models, GNNs, and multimodal generative approaches~\citep{ngiam2021scene,zhou2022hivt,tang2024hpnet}.
While these methods achieve strong benchmark performance, they rely heavily on large-scale datasets~\citep{caesar2020nuscenes,chang2019argoverse,wilson2023argoverse,zhan2019interaction,sun2020scalability}, making the underlying data distribution a critical factor in determining model behavior.
A key characteristic of these datasets is a pronounced long-tail imbalance: common low-density scenarios dominate, while complex interaction-heavy scenarios are rare~\citep{makansi2021exposing,wang2023fend,pourkeshavarz2023learn,chen2024criteria}.
As a result, model performance evaluated on global averages often masks substantial degradation in dense, safety-critical situations.
Recent work has explored long-tail robustness through debiased or contrastive representation learning~\citep{wang2023fend,zhang2024tract,zhou2022long,lan2024hi}, but these approaches primarily operate at the feature level.
They do not explicitly address the scenario-level imbalance or consider how individual samples contribute to the overall distribution.
In contrast, our method examines the importance of each scenario and applies a density-aware selection strategy that reshapes the dataset at the sample level.
By constructing a compact yet balanced subset, it improves coverage of rare high-density scenarios and enables more robust prediction in complex settings.

\noindent {\bf Training Sample Selection.} 
Deep neural networks, especially Transformers, typically require large-scale training data and incur substantial computational cost. To improve data efficiency, prior work has explored strategies such as frequent parameter updates~\citep{robbins1951stochastic}, reducing the number of effective iterations~\citep{sutskever2013importance}, and adaptive learning rate schedules~\citep{kingma2014adam,duchi2011adaptive}. A more direct line of work reduces the training data volume itself. Dataset condensation compresses raw datasets into compact synthetic samples~\citep{wang2018dataset,zhao2020dataset,zhao2021dataset,kim2022dataset,wang2022cafe,zhao2023dataset,cazenavette2022dataset}, while coreset selection constructs representative subsets that approximate the original data distribution~\citep{har2004coresets,coleman2019selection,margatina2021active,mirzasoleiman2020coresets}. More generally, subset selection has been studied through submodular optimization and active learning, with later work improving scalability and embedding-based representativeness and diversity~\citep{wei2015submodularity,malioutov2016large,qian2024sub}.
Despite their success, existing dataset condensation~\citep{nguyen2020dataset,loo2022efficient,zhou2022dataset,sajedi2023datadam,zhao2023improved} and subset selection~\citep{killamsetty2021grad,paul2021deep} are largely developed for image classification or generic annotation settings. They do not directly address the unique challenges of trajectory prediction, where each sample involves multi-agent temporal dynamics, interaction structure, and density-conditioned difficulty.
In contrast, our framework is tailored to trajectory prediction datasets and performs density-aware, gradient-informed sample selection at the scenario level, enabling compact subset construction while preserving coverage of rare high-density scenarios.

\section{Method}

Our goal is to address the pronounced long-tail imbalance in scenario density that biases training toward abundant low-density scenarios and leads to systematic undertraining in high-density cases.
Driven by this motivation, we propose a data-centric framework Den-TP to curate a compact and density-balanced subset which enables models trained on it to maintain overall accuracy while improving robustness in complex, high-density scenarios.

\noindent \textbf{Preliminary.}
Denote the training set as $\mathcal{D}=\{S_{j}\}_{j=1}^{N}$. %
Each sample is represented as a triple $S=(X, Y, \mathcal{O})$, where $X$ and $Y$ denote the observed trajectories and future trajectories of all agents, and $\mathcal{O}$ denotes the driving context information (e.g., maps). 
The goal is to estimate $Y$ conditioned on $X$ and $\mathcal{O}$. 
We aim to construct a compact subset $\mathcal{C}\subseteq\mathcal{D}$ of target size $B$. A model trained on $\mathcal{C}$ is expected to achieve trajectory prediction performance comparable to that obtained from training on the full dataset $\mathcal{D}$. Our method is formally described in Algorithm \cref{alg:algo}. Next, we explain the key algorithmic components in our method.

\vspace{0.5em}
\begin{algorithm}[!t]
\caption{\footnotesize Sample Selection for Trajectory Prediction}
\small
\label{alg:algo}
\textbf{Input:} 
Full Dataset $\mathcal{D}$, 
interval $\tau$, 
ratio $\alpha$, 
submodular function $P(\cdot)$\\
\textbf{Output:} Target dataset $\mathcal{C}$ 

\begin{algorithmic}[1]
\State Initialize: $\mathcal{C} \gets \emptyset$;
\State Initialize: $B \gets \lfloor \alpha |\mathcal{D}| \rfloor$; \Comment{Set target size}
\State Partitioning: $\mathcal{D}_{k}$;

\For{$k \in \{K, K-1, ..., 1\}$:} \Comment{Reverse order}
    \State $\mathcal{C}_k \gets \emptyset$;
    \State $n_{k} \gets \texttt{DynamicSelect}(B, k)$;
    \If{$n_{k} \ge |\mathcal{D}_k|$}: \Comment{Budget covers this partition}
        \State $\mathcal{C}_k \gets \mathcal{D}_k$; \Comment{Keep all samples}
        \State $\mathcal{C} \gets \mathcal{C} \cup \mathcal{C}_k$;
    \Else: \Comment{Compute gradient features for samples in $\mathcal{D}_k$}
        \For{$n = 1$ to $n_{k}$:} \Comment{Iterate $n_{k}$ times}
            \State $S_j \gets \arg\min_{S_j \in \mathcal{D}_k \setminus \mathcal{C}_{k}} P(S_j)$;
            \State $\mathcal{C}_{k} \gets \mathcal{C}_{k} \cup \{S_j\}$;
        \EndFor
        \State $\mathcal{C} \gets \mathcal{C} \cup \mathcal{C}_k$;
    \EndIf
    \State $B \gets B - |\mathcal{C}_k|$;  \Comment{Update remaining size}
\EndFor

\State \textbf{Return} \(\mathcal{C}\)

\vskip 0.5\baselineskip 

\Function{DynamicSelect}{$B, k$}:
    \State \Return $\min\!\left(|\mathcal{D}_k|, \left\lfloor \frac{B}{k} \right\rfloor\right)$
\EndFunction
\end{algorithmic}
\end{algorithm}

\noindent\textbf{Data Partitioning.}
A central characteristic of trajectory prediction datasets is the long-tail imbalance in scenario density: low-density scenarios dominate, while high-density scenarios with complex multi-agent interactions are comparatively rare.
This uneven distribution limits the dataset’s coverage of dense, safety-critical situations and biases gradient updates during training.
Without density-based partitioning, gradient updates are dominated by abundant low-density samples, biasing the model toward sparse interactions and leading to systematic undertraining on high-density cases.
Consequently, models receive insufficient signal to learn interaction patterns that generalize to dense scenarios.
Explicitly partitioning the dataset by scenario density provides a principled first step toward mitigating this imbalance.
Specifically, we first compute a density level $\rho(S_{j})$ for each sample $S_{j} \in \mathcal{D}$ based on the number of agents present. Then, we use a fixed interval $\tau$ to partition the dataset into $K$ disjoint subsets $\mathcal{D}_k$ for $k\in [K]$ based on these density values such that $S\in \mathcal{D}_k$ if $\rho(S)\in [\rho_{\min} + (k-1)\tau,\;\rho_{\min} + k\tau )$, where $\rho_{\min}$ is the minimum density level in $\mathcal{D}$.

\noindent\textbf{Gradient Extraction.}
Given an input sample, the trajectory predictor outputs a multimodal future prediction tensor
$\hat{\mathbf{Y}} \in \mathbb{R}^{F \times M \times H \times D}$ where $F$ is the number of prediction modes, $M$ is the number of agents,
$H$ is the prediction horizon, and $D$ is the trajectory output dimension.
We calculate the total loss as $\mathcal{L}=\mathcal{L}_{\text{reg}}+\mathcal{L}_{\text{cls}}$,
where $\mathcal{L}_{\text{reg}}$ is a negative log-likelihood regression loss computed on the best-matched prediction mode under the variety loss formulation, and $\mathcal{L}_{\text{cls}}$ is a cross-entropy loss used to optimize the predicted mode probabilities.
By backpropagating the total loss, we extract gradient features associated with the prediction outputs:
\begin{equation}
    \mathbf{G} = \nabla_{\hat{\mathbf{Y}}}\,\mathcal{L},
    \label{eq:grad}
\end{equation}
where $\mathbf{G}\in\mathbb{R}^{F \times M \times H \times D}$ has the same shape as $\hat{\mathbf{Y}}$.
Let $\mathbf{E} \in \mathbb{R}^{F \times M \times d}$ denote the corresponding decoder latent tensor, where $d$ is the decoder embedding dimension.
Since $\mathbf{G}$ and $\mathbf{E}$ have different tensor shapes, we apply a shape-alignment operator $\phi(\cdot)$, implemented via flattening and padding, to map them into a common vector space. The fused gradient feature is then defined as:
\begin{equation}
\mathbf{g} =
\phi(\mathbf{G}) \odot \phi(\mathbf{E}),
   \label{eq:joint}
\end{equation}
where $\odot$ denotes element-wise multiplication. Thus, $\mathbf{g}$ captures joint information from both loss sensitivity and decoder representations.

\noindent \textbf{Sample Selection.}
Recall that $B$ is the target size of the constructed training subset. 
To ensure fair representation across scenario complexities, we adopt a dynamic allocation strategy.
For each partition $\mathcal{D}_k$, a local budget $n_k$ is assigned by the \texttt{DynamicSelect} function (see Algorithm~\ref{alg:algo}). 
This mechanism prioritizes high-density subsets that naturally contain fewer samples, preventing them from being underrepresented when the global budget is small. Formally, the allocation satisfies:
\begin{equation}
    \sum_{k=1}^{K} n_k = B, \quad n_k \geq 0.
    \label{eq:budget_constraint}
\end{equation}
Given the $K$ disjoint subsets and their corresponding gradient feature sets $\mathcal{G}_k$, we define a submodular score function $P(\cdot)$ to evaluate candidate samples as:
\begin{equation}
    P(S_{j}) = \sum_{S_i \in \mathcal{C}_{k}} \frac{\mathbf{g}_i \cdot \mathbf{g}_j}{\|\mathbf{g}_i\| \|\mathbf{g}_j\|} - \sum_{S_i \in \mathcal{D}_k \setminus \mathcal{C}_{k}} \frac{\mathbf{g}_i \cdot \mathbf{g}_j}{\|\mathbf{g}_i\| \|\mathbf{g}_j\|},
    \label{eq:submodular}
\end{equation}
where cosine similarity is used to measure the affinity between sample $S_j$ and other samples in the same subset.
Minimizing $P(S_j)$ favors samples that are less redundant with the current selection set $\mathcal{C}_k$ while remaining representative of the unselected portion $\mathcal{D}_k \setminus \mathcal{C}_k$.
We then apply a greedy optimization strategy, iteratively selecting the sample as follows:
\begin{equation}
    S^* = \arg\min_{S_j \in \mathcal{D}_k \setminus \mathcal{C}_{k}} P(S_j).
    \label{eq:argmin}
\end{equation}
At each iteration, the selected sample $S^{*}$ is added to $\mathcal{C}_{k}$. This process continues
until the number of selected samples reaches the budget $n_{k}$ for subset $\mathcal{D}_k$. 
Notably, we process the subsets starting with those having higher density levels, as these subsets tend to be scarce and underrepresented. This prioritization is consistent with our dynamic allocation strategy, which ensures that complex, high-density scenarios are not discarded in early pruning and that long-tailed scenarios remain adequately represented. For subsets $\mathcal{D}_k$ where $|\mathcal{D}_k| \leq n_k$, we directly set $\mathcal{C}_k = \mathcal{D}_k$.
Finally, the target subset is obtained as $\mathcal{C} = \bigcup_{k=1}^{K}\mathcal{C}_{k}$.
By incorporating gradient-based similarity into submodular greedy selection, our method maximizes coverage of the gradient space while maintaining diversity, producing a smaller subset that remains representative and informative. 
In addition, we account for efficiency: the additional cost of sample selection is dominated by gradient feature extraction and greedy updates. Notably, for partitions with $n_k \ge |\mathcal{D}_k|$, we directly set $\mathcal{C}_k=\mathcal{D}_k$ and skip gradient-based selection. Let $\mathcal{I}=\{k\in[K]\mid n_k<|\mathcal{D}_k|\}$ denote the set of partitions that require greedy selection, and let $d$ be the feature dimension. The overall complexity can be approximated as $\mathcal{O}(\text{selection}) \approx \sum_{k\in\mathcal{I}}\mathcal{O}\!\left(|\mathcal{D}_k|\cdot d\right) + \sum_{k\in\mathcal{I}}\mathcal{O}\!\left(n_k\,|\mathcal{D}_k|\cdot d\right)$, where the first term corresponds to gradient feature extraction for samples in partitions that require greedy selection, and the second term corresponds to greedy updates within these partitions.

\noindent \textbf{Why Naive Strategies Fall Short.}  
Several alternative strategies can be used to modify the training data, but they do not adequately address the underlying distributional imbalance under a fixed data budget.
\textbf{\textit{Re-weighting}} adjusts loss weights for high-density samples while keeping the original dataset intact.
\textbf{\textit{Augmenting}} duplicates scarce high-density cases to increase their frequency.
\textbf{\textit{High-density+Random}} retains all high-density samples and fills the remaining budget with random selections.
\textbf{\textit{Epoch-wise}} resamples a subset of data at the beginning of each training epoch.
In contrast, Den-TP focuses on \emph{curating an existing long-tailed dataset under a fixed budget} to achieve a better accuracy--cost trade-off by controlling both representativeness and redundancy.

\begin{table}[t]
    \centering
    \small
    \setlength{\tabcolsep}{2.5pt}
    \renewcommand{\arraystretch}{0.98}
    \begin{tabular}{l|c|c|c|c}
        \toprule
        Method & \#Samples & minADE$\downarrow$ & minFDE$\downarrow$ & MR$\downarrow$ \\
        \midrule
        Augmenting   & 220k & 0.718 & 1.106 & 0.115 \\
        Weighting    & 190k & 0.715 & 1.108 & 0.114 \\
        Epoch-wise   & 95k  & 0.752 & 1.189 & 0.130 \\
        Random       & 95k  & 0.750 & 1.175 & 0.126 \\
        High-density+Random & 95k & 0.724 & 1.111 & 0.117 \\
        \rowcolor{RowColor}
        \textbf{Den-TP} & 95k & \textbf{0.706} & \textbf{1.074} & \textbf{0.110} \\
        \bottomrule
    \end{tabular}
        \caption{\footnotesize Comparison with different selection strategies on Argoverse 1. All methods are evaluated at 50\% subset unless otherwise specified. Den-TP outperforms all baselines across minADE, minFDE, and MR.}
    \label{tab:add_baseline}
    \vspace{1em}
\end{table}

Although these approaches alter sampling frequencies, they do not reshape the data distribution in a principled manner.
Random down-sampling and epoch-wise re-selection reduce dataset size but either discard informative samples or introduce instability (Table~\ref{tab:add_baseline}, lines 3–4).
Re-weighting or duplication increases the presence of high-density samples but often amplifies redundancy rather than improving coverage of diverse interaction patterns (lines 1–2).
These observations show that effective density balancing requires controlling both representativeness and redundancy, not merely adjusting sample counts.
Preserving all high-density scenarios is intuitively beneficial, but without a principled way to control redundancy and ensure representativeness, the gains remain limited (line 5).
High-density cases are especially important in trajectory prediction, as they contain complex multi-agent interactions that are not adequately captured by oversampling or naive mixing.
A key empirical observation is that competence learned from dense, interaction-rich scenarios tends to transfer better to simpler ones (Figure~\ref{fig:fig2}), whereas the reverse transfer is much weaker.
This asymmetry motivates Den-TP to preserve scarce high-density scenarios while avoiding redundant low-density ones.

\section{Experiments}

\label{sec:experiments}
\begin{table*}[t]
    \centering %
    \renewcommand{\arraystretch}{0.95} %
    \setlength{\tabcolsep}{4pt} %
    \begin{tabular}{l | c | ccc | ccc | ccc}
        \toprule
        \multicolumn{1}{l}{\multirow{2}{*}{Methods}} 
        & \multicolumn{1}{c}{\multirow{2}{*}{$\alpha$(\%)}}
        & \multicolumn{3}{c}{HiVT-64} 
        & \multicolumn{3}{c}{HiVT-128} 
        & \multicolumn{3}{c}{HPNet} \\
        \cmidrule(lr){3-5} \cmidrule(lr){6-8} \cmidrule(lr){9-11}
        \multicolumn{1}{c}{} & \multicolumn{1}{c}{} 
        & minADE$\downarrow$ & minFDE$\downarrow$ & \multicolumn{1}{c}{\phantom{x}MR$\downarrow$\phantom{x}}
        & minADE$\downarrow$ & minFDE$\downarrow$ & \multicolumn{1}{c}{\phantom{x}MR$\downarrow$\phantom{x}} 
        & minADE$\downarrow$ & minFDE$\downarrow$ & \phantom{x}MR$\downarrow$\phantom{x} \\
        \midrule
        Argoverse 1 & 100
        & \textbf{0.695} & \textbf{1.037} & \textbf{0.109} 
        & \textbf{0.666} & \textbf{0.978} & \textbf{0.091} 
        & \textbf{0.647} & \textbf{0.871} & \textbf{0.070} \\
        \midrule
        Random  &  \multirow{4}{*}{60}
                & 0.745 & 1.163 & 0.132 
                & 0.719 & 1.078 & 0.129  
                & 0.680 & 0.951 & 0.091  \\
        Cluster & 
                & 0.716 & 1.097 & 0.121 
                & 0.697 & 1.025 & 0.108
                & 0.673 & 0.930 & 0.081 \\
        Herding &
                & 0.723 & 1.101 & 0.125 
                & 0.685 & 1.018 & 0.106
                & 0.666 & 0.922 & 0.085 \\
        \rowcolor{RowColor}
        \textbf{Den-TP}  &
                &\textbf{0.702} &\textbf{1.064} &\textbf{0.110}                
                &\textbf{0.674} &\textbf{0.994} &\textbf{0.093}                
                &\textbf{0.653} &\textbf{0.901} &\textbf{0.071} \\
        \midrule
        Random  & \multirow{4}{*}{50}
                & 0.750 & 1.175 & 0.137 
                & 0.728 & 1.098 & 0.126 
                & 0.687 & 0.967 & 0.091 \\
                
        Cluster & 
                & 0.725 & 1.117 & 0.124 
                & 0.692 & 1.033 & 0.118 
                & 0.676 & 0.952 & 0.085 \\
                
        Herding &
                & 0.728 & 1.107 & 0.126 
                & 0.698 & 1.036 & 0.119 
                & 0.674 & 0.938 & 0.089 \\
        \rowcolor{RowColor}
        \textbf{Den-TP}  &
                & \textbf{0.706} & \textbf{1.074} & \textbf{0.110}                
                & \textbf{0.684} & \textbf{1.022} & \textbf{0.101}                
                & \textbf{0.661} & \textbf{0.913} & \textbf{0.074} \\
        \midrule
        Random  & \multirow{4}{*}{40}
                & 0.752 & 1.183 & 0.139 
                & 0.727 & 1.109 & 0.126 
                & 0.696 & 0.987 & 0.099\\
        Cluster & 
                & 0.732 & 1.141 & 0.127 
                & 0.703 & 1.058 & 0.121 
                & 0.681 & 0.962 & 0.089 \\ 
        Herding &
                & 0.722 & 1.123 & 0.128
                & 0.704 & 1.056 & 0.119 
                & 0.684 & 0.956 & 0.093\\
        \rowcolor{RowColor}
        \textbf{Den-TP}  &
                &\textbf{0.711} &\textbf{1.088} &\textbf{0.114} 
                & \textbf{0.696} & \textbf{1.048} & \textbf{0.106} 
                & \textbf{0.671} & \textbf{0.931} & \textbf{0.076}  \\
        \bottomrule    
    \end{tabular}
    \caption{\footnotesize Performance comparison results on Argoverse 1 with data retention ratios of 60\%, 50\%, and 40\%. The compared methods include Random Selection, K-Means Clustering, and Herding Selection. The model used for data selection is HiVT-64, while the evaluation is conducted on HiVT-64, HiVT-128, and HPNet. $\alpha$ (\%) represents the proportion of retained data relative to the full training set.}
    \label{tab:argo1_result}
    \vspace{1em}
\end{table*}

\subsection{Benchmarks and Setup}
\noindent\textbf{Datasets.} 
We evaluated Den-TP on Argoverse Motion Forecasting Dataset 1.1~\citep{chang2019argoverse} and Argoverse 2~\citep{wilson2023argoverse}. The Argoverse 1 dataset contains 323,557 real-world driving scenarios. All the training and validation scenarios are 5-second sequences sampled at 10 Hz. The length of the historical trajectory for each scenario is 2 seconds, and the length of the predicted future trajectory is 3 seconds. The Argoverse 2 dataset contains 250,000 scenarios, with the same sampling frequency of 10 Hz. Each scenario has a longer observation window of 5 seconds and a longer prediction horizon of 6 seconds.

\noindent\textbf{Baselines.}
For Argoverse 1, we evaluate Den-TP on two models HiVT~\citep{zhou2022hivt} and HPNet~\citep{tang2024hpnet} for evaluation. 
For Argoverse 2, we evaluate Den-TP using QCNet~\citep{zhou2023query} and DeMo~\citep{zhang2025decoupling}. For a more comprehensive comparison, we also include the following three data selection baselines: 
\noindent(1) Random Selection~\citep{rebuffi2017icarl}: randomly selects a target proportion of training samples from the original dataset.
\noindent(2) K-Means Clustering~\citep{likas2003global}: clusters trajectories within the observation window based on their features, and then selects from each cluster the trajectory sample closest to the cluster center as a representative.
\noindent(3) Herding Selection~\citep{castro2018end}: a greedy strategy that first computes the mean feature of all trajectories within the observation window and then iteratively selects trajectory samples that bring the mean of the selected subset as close as possible to the overall mean.

\noindent\textbf{Metrics.}
Following the baseline settings, each model predicts a total of 6 future trajectories. We evaluate prediction performance using minimum Average Displacement Error (minADE), minimum Final Displacement Error (minFDE), and Missing Rate (MR).

\noindent\textbf{Implementation Details.}
We primarily use pre-trained HiVT-64 and QCNet as backbone models to perform sample selection on Argoverse 1 and Argoverse 2, respectively. To evaluate the performance of the selected subset, we follow their official training and validation protocols. We experimented with different selection ratios, different intervals, and assessed the prediction accuracy of the trajectory models after training on the corresponding subsets.

\subsection{Main Results}
We present the primary experimental results in this section. 
Table~\ref{tab:argo1_result} shows the strong performance of our selected subset on the Argoverse 1 dataset across all compression rates. Following the same experimental setup as the baseline models, we trained HiVT and HPNet from scratch on the subset. Our method substantially reduces data volume while maintaining comparable performance. Even with only half of the data, models trained on our selected subset still perform comparably to those trained on the full dataset and consistently outperform random selection, clustering, and herding. Furthermore, our subset also demonstrates strong results on larger models such as HiVT-128 and HPNet.

\begin{table*}[t] %
    \centering
    \renewcommand{\arraystretch}{0.95} %
    \setlength{\tabcolsep}{7pt} %
    \begin{tabular}{l|c|ccc|ccc}
        \toprule
        \multicolumn{1}{l}{\multirow{2}{*}{Method}} 
        & \multicolumn{1}{c}{\multirow{2}{*}{$\alpha$(\%)}}
        & \multicolumn{3}{c}{QCNet} 
        & \multicolumn{3}{c}{DeMo} \\
        \cmidrule(lr){3-5} \cmidrule(lr){6-8}
        \multicolumn{1}{c}{}&  \multicolumn{1}{c}{}
        & minADE$\downarrow$ & minFDE$\downarrow$ & \multicolumn{1}{c}{\phantom{x}MR$\downarrow$\phantom{x}} 
        & minADE$\downarrow$ & minFDE$\downarrow$ & \phantom{x}MR$\downarrow$\phantom{x}  \\
        \midrule
        Argoverse 2
            & 100
            & \textbf{0.724} & \textbf{1.258} & \textbf{0.162} 
            & \textbf{0.657} & \textbf{1.254} & \textbf{0.163}  \\
        \midrule
        Random  
            & \multirow{4}{*}{60}
            & 0.787 & 1.419 & 0.208
            & 0.755 & 1.433 & 0.198  \\
        Cluster 
            & 
            & 0.773 & 1.406 & 0.192 
            & 0.693 & 1.386 & 0.187  \\
        Herding 
            & 
            & 0.778 & 1.402 & 0.191 
            & 0.701 & 1.494 & 0.189  \\
        \rowcolor{RowColor}
        \textbf{Den-TP} 
            & 
            & \textbf{0.740} & \textbf{1.316} & \textbf{0.163} 
            & \textbf{0.682} & \textbf{1.344} & \textbf{0.164}  \\
        \midrule
        Random  
            & \multirow{4}{*}{50}
            & 0.805 & 1.447 & 0.219 
            & 0.756 & 1.448 & 0.203  \\
        Cluster 
            & 
            & 0.798 & 1.435 & 0.193 
            & 0.732 & 1.437 & 0.191  \\
        Herding 
            & 
            & 0.782 & 1.407 & 0.190 
            & 0.731 & 1.434 & 0.190  \\
        \rowcolor{RowColor}
        \textbf{Den-TP}  
            & 
            & \textbf{0.754} & \textbf{1.352} & \textbf{0.172} 
            & \textbf{0.704} & \textbf{1.414} & \textbf{0.173}  \\
        \midrule
        Random  
            & \multirow{4}{*}{40}
            &0.811 &1.471 &0.226 
            &0.763 &1.475 &0.202 \\
        Cluster 
            & 
            &0.813 &1.495 &0.214 
            &0.732 &1.456 &0.195 \\
        Herding 
            & 
            & 0.807 & 1.454 & 0.198 
            & 0.739 & 1.460 & 0.197  \\
        \rowcolor{RowColor}
        \textbf{Den-TP}  
            & 
            &\textbf{0.778} &\textbf{1.410} &\textbf{0.183} 
            & \textbf{0.723} &\textbf{1.450} & \textbf{0.191} \\
        \bottomrule
    \end{tabular}
    \caption{\footnotesize Performance comparison results on Argoverse 2 with data retention ratios of 60\%, 50\%, and 40\%. The compared methods include Random Selection and K-means clustering. The model used for data selection is QCNet, while the evaluation is conducted on QCNet and DeMo.}
    \label{tab:argo2}
    \vspace{1em}
\end{table*}

\begin{figure*}[t]
\centering
\includegraphics[width=\linewidth]{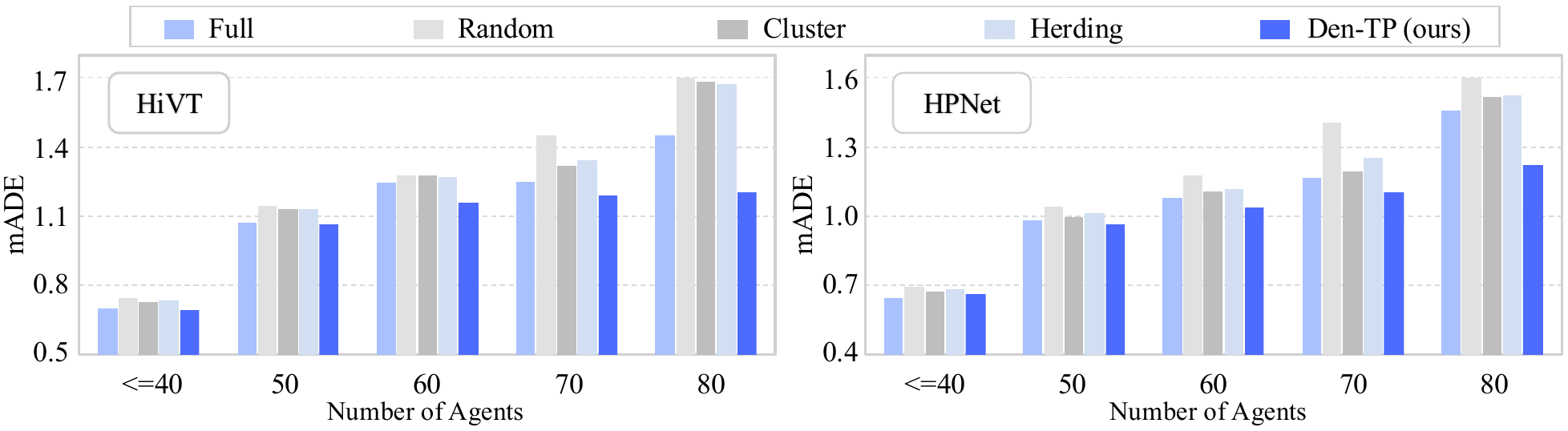}

\caption{\footnotesize Comparison of model performance across scenario densities when trained on the full dataset, a 50\% random subset, and a 50\% Den-TP subset, where Den-TP consistently achieves superior results. Each bar group represents a different density level, and bar height indicates the corresponding minADE. Prediction error increases with density for all methods, but Den-TP consistently achieves lower minADE in high-density scenarios.}
\label{fig:table3}
\vspace{1em}
\end{figure*}

\noindent\textbf{Performance Enhancement.} 
Scenario density in autonomous driving varies substantially, yet most existing trajectory prediction datasets are dominated by low-density scenarios. From a safety perspective, however, an ideal trajectory predictor must perform reliably across the full spectrum of scenario complexities. To this end, we evaluate our proposed method on multiple models and across different density levels.
As shown in 
Figure~\ref{fig:table3}, our method consistently outperforms models trained on the full dataset, particularly in high-density scenarios, on both the Argoverse 1 and Argoverse 2 datasets. In low-density settings (fewer than 40 agents), our selected subset achieves performance nearly identical to the full dataset, with only marginal increases of minADE and minFDE, while MR remains almost unchanged, which is negligible given the limited interactions in such scenarios.  

When the agent density increases, our method brings the most substantial gains. Compared to all other baselines, Den-TP achieves lower displacement errors and notably reduces MR. For example, when the number of agents exceeds 80, Den-TP cuts the missing rate by more than 8\% relative to random selection, and also outperforms clustering and herding by clear margins.  
These results highlight that while clustering and herding provide partial improvements by ensuring representativeness, they are still insufficient for handling highly complex traffic scenarios. In contrast, Den-TP effectively balances scenario density while selecting informative samples, leading to consistently superior performance across all density levels. Similarly, HPNet trained on our selected subset even outperforms its counterpart trained on the full dataset.

\begin{table*}[htbp]
    \centering
    \setlength{\tabcolsep}{6 pt} %
    \renewcommand{\arraystretch}{0.95} %
    \begin{tabular}{l | cc | ccc}
        \toprule
        \multicolumn{1}{l}{\multirow{2}{*}{Variants}} 
        & \multicolumn{2}{c}{Data Distribution(\%)} 
        & \multicolumn{3}{c}{Model Performance} \\
        \cmidrule(lr){2-3} \cmidrule(lr){4-6} 
        \multicolumn{1}{c}{}
        & Agent$<$40 & \multicolumn{1}{c}{Agent$>=$40} 
        & minADE$\downarrow$ & minFDE$\downarrow$ & \phantom{x}MR$\downarrow$\phantom{x} \\
    \midrule
    Full dataset 
        &85.17 &14.83
        &0.695 &1.037 &0.109 \\ 
    \midrule
    Random 
        &85.16 &14.84
        &0.750 & 1.175 & 0.137 \\ 
    Den-TP w/ Submodular 
        &85.17 &14.83
        &0.724 &1.115 &0.116 \\ 
    Den-TP w/ Partition 
        &70.35 &29.65
        &0.729 &1.116 &0.118 \\ 
    \rowcolor{RowColor}
    \textbf{Den-TP} 
        &70.35 &29.65
        &\textbf{0.706} &\textbf{1.074} &\textbf{0.110} \\ 
    \bottomrule
    \end{tabular}
    \caption{\footnotesize Performance comparison of data selection strategies on HiVT trained with Argoverse 1. This table shows the impact of partitioning and selection with Submodular Gain strategies on data distribution and model performance. The whole dataset and random selection serve as baselines, while different variations of Den-TP are evaluated. Our method (Den-TP) integrates both strategies, achieves the best results by maintaining a balanced data distribution and reducing minADE, minFDE, and MR.}
    \label{tab:ablation}
    \vspace{1em}
\end{table*}

\begin{figure*}[t]
\centering
\begin{minipage}{0.49\linewidth}
    \centering
    \includegraphics[width=\linewidth]{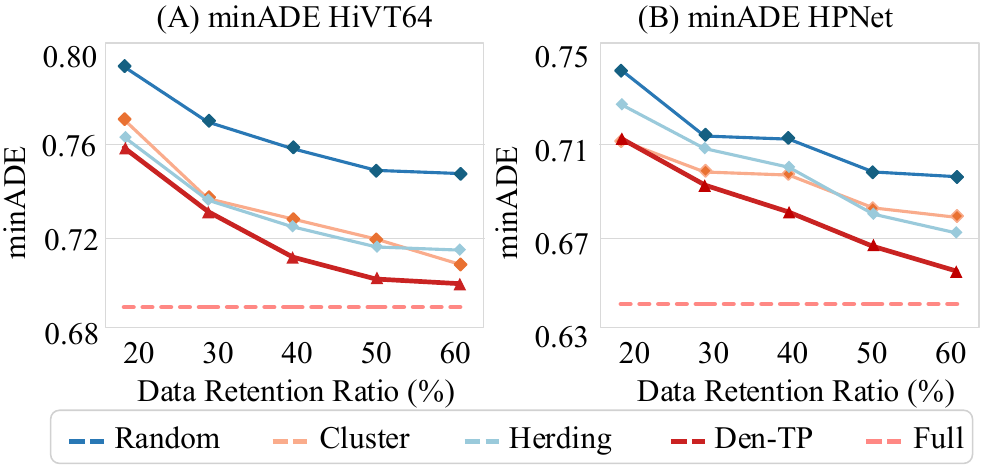}
    \caption{\footnotesize Performance on Argoverse1 for HiVT-64 (A) and HPNet (B) under different data retention ratios. Models are trained using Random Selection, K-means Clustering, Herding, and our Den-TP, with the full-dataset performance shown as a reference (dashed line). }
    \label{fig:ratio}
    \vspace{1em}
\end{minipage}
\hfill
\begin{minipage}{0.48\linewidth}
    \centering
    \includegraphics[width=\linewidth]{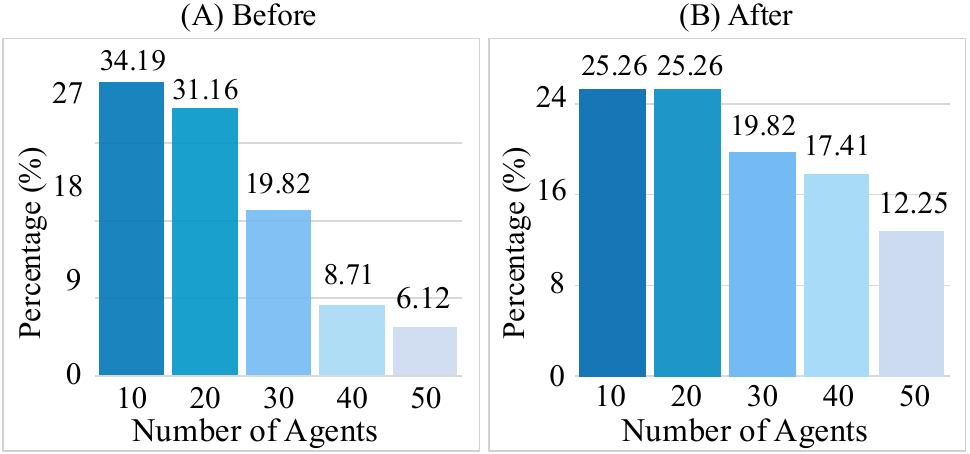}
    \caption{\footnotesize Distribution of scenarios categorized by agent density in Argoverse 1 before (\textbf{left}) and after (\textbf{right}) applying Den-TP with 50\% data retention ratio. The original dataset (A) is heavily skewed toward low-density scenarios, whereas Den-TP (B) produces a more balanced distribution by preserving a higher proportion of medium- and high-density scenarios.}
    \label{fig:balance}
    \vspace{1em}
\end{minipage}
\end{figure*}

\noindent\textbf{Generalization Across Datasets.} 
We further evaluated our proposed method on the Argoverse 2 dataset, which presents greater challenges due to its more diverse driving scenarios and longer prediction horizons. As shown in Table~\ref{tab:argo2}, our method consistently outperforms other data selection strategies across all data retention rates, achieving lower minADE and minFDE while maintaining a lower MR. 
These results further validate the robustness of our approach, as it maintains strong performance across different datasets. This demonstrates that our method is not only effective within a specific dataset but also generalizes well to more complex and diverse trajectory scenarios, such as those found in Argoverse 2.

\noindent \textbf{Data Retention Ratio.}
To examine the impact of different data retention ratios on model performance, we conducted experiments with retention rates $\alpha \in \{60, 50, 40, 30, 20\}\%$ as shown in Figure~\ref{fig:ratio}. When higher model performance is required, retaining 50\% of the data already achieves results comparable to training with the full dataset. This demonstrates the effectiveness of our Den-TP method, as the selected subset is of higher quality compared to equally sized subsets chosen by other methods. Furthermore, under limited computational resources, even retaining only 20\% of the data still yields reasonably good results.

\noindent\textbf{Density Balancing.} Our method explicitly controls scenario density distribution during selection, ensuring a more balanced dataset, as illustrated in Figure~\ref{fig:balance}. In contrast, random selection fails to maintain this balance, leading to uneven representation of scenarios with varying complexity and weaker generalization. As shown in Table~\ref{tab:ablation} line 4, applying scenario balancing alone already improves performance compared to random selection, demonstrating that controlling scenario density enhances the effectiveness of trajectory prediction models. However, balancing alone remains insufficient. By further integrating submodular selection to account for sample informativeness, our method achieves the best performance across all metrics. These findings indicate that while scenario balancing is beneficial, it is insufficient to achieve optimal performance without also considering sample informativeness.

\begin{figure}[t]
    \centering
    \includegraphics[width=1\linewidth]{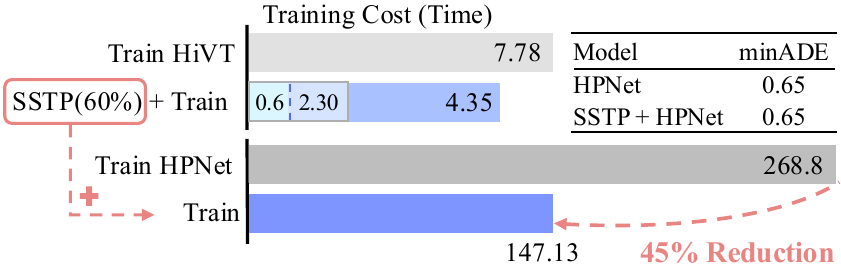}
    \caption{Training cost comparison between full-dataset HPNet training and the proposed Den-TP pipeline. Using Den-TP to preselect 60\% of the data with HiVT reduces the subsequent HPNet training time by 45\%, while maintaining the same minADE.}
    \label{fig:time_saving}
    \vspace{1em}
\end{figure}

\noindent\textbf{Efficiency.} Our method significantly reduces computational time while maintaining strong performance, as shown in Figure \ref{fig:time_saving}. Training the HiVT model on the full dataset requires 7.78 hours. In contrast, utilizing our Den-TP method to select a 60\% subset requires only 0.6 hour for pre-training and 2.30 hours for selection,  reducing the overall training time. When training on the selected subset, the total training time decreases to 7.25 hours (0.6 + 2.30 + 4.35), with only a minor increase of 0.007 in minADE.

Beyond efficiency on the original backbone, a key advantage of Den-TP is that the selected subset is model-agnostic and can be directly reused to train other trajectory predictors. For instance, when applied to HPNet, training on the full dataset takes 307.2 hours, whereas training with the selected subset reduces the time by over 100 hours, cutting the training cost by nearly 45\%, while achieving comparable or even better accuracy.
These results highlight the efficiency of our approach in balancing training cost and model performance.

\noindent\textbf{Generalizability across Backbones.} To further assess the generalizability of our method, we evaluated its performance with different backbone models. Specifically, we used pre-trained HiVT-64 and HPNet as feature extractors on the Argoverse 1 dataset and conducted experiments under varying data retention ratios. As shown, regardless of the backbone used for subset selection, the final trajectory prediction performance remains nearly identical. This demonstrates that our subset selection strategy is largely backbone-agnostic, underscoring its robustness and broad applicability for optimizing diverse trajectory prediction models.

\noindent\textbf{Safety-critical evaluation.}
We further report JADE/JFDE~\cite{weng2023joint}, NLL, and collision rate on the full validation set and a high-density subset. Table~\ref{tab:safety_metrics} shows that our method consistently reduces collision rate and achieves the best JADE/JFDE on the safety-critical split, demonstrating improved performance in challenging, high-interaction scenarios.

\begin{table}[t]
\centering
\footnotesize
\setlength{\tabcolsep}{1.5 pt}
\renewcommand{\arraystretch}{1.1}
\begin{tabular}{l | c c c | c c c}
    \toprule
    & \multicolumn{3}{c|}{Full Eval} & \multicolumn{3}{c}{High-Density Eval} \\
    \midrule
    Method & JADE/JFDE$\downarrow$ & NLL$\downarrow$ & CR$_{\text{mean}}\downarrow$
           & JADE/JFDE$\downarrow$ & NLL$\downarrow$ & CR$_{\text{mean}}\downarrow$ \\
    \midrule
    Ours   & 0.819/1.684 & 0.030 & \textbf{0.045}
           & \textbf{1.677/2.179} & \textbf{0.069} & \textbf{0.125} \\
    Full   & \textbf{0.816/1.669} & \textbf{0.028} & 0.046
           & 1.741/2.381 & 0.074 & 0.134 \\
    Random & 0.862/1.761 & 0.151 & 0.049
           & 1.968/2.774 & 0.234 & 0.148 \\
    \bottomrule
\end{tabular}
\caption{
Safety-related metrics on the full validation set and the high-density subset. Lower is better for all metrics.
}
\label{tab:safety_metrics}
\vspace{1em}
\end{table}

\section{Conclusion}
\label{sec:conclusion}
We presented Den-TP, a data-centric framework for dataset curation and evaluation in trajectory prediction.  
By explicitly accounting for the long-tail imbalance in scenario density, Den-TP constructs compact yet representative subsets that reduce training cost while preserving overall accuracy and improving robustness in high-density, interaction-heavy scenarios.  
Our density-conditioned evaluation further exposes failure modes obscured by standard aggregate metrics, showing that balanced coverage across density regimes is essential for robust prediction.  
These results highlight the importance of data-centric curation for scalable trajectory prediction and motivate future work on richer density indicators and broader applications in autonomous driving.

\noindent\textbf{Limitations.}
Our method still incurs non-negligible computational overhead during gradient extraction and submodular optimization. Although this is a one-time cost, reducing this overhead would improve practicality. Moreover, performance degrades under extremely low data retention ratios (e.g., 10\%), highlighting the difficulty of preserving robustness with very limited data. Finally, our study is restricted to trajectory prediction on the Argoverse benchmarks; extending Den-TP to other autonomous driving tasks and datasets remains an important direction for future work.

\section*{Acknowledgement}
This work is supported in part by NSF CAREER CCF-2340482 and Sony Faculty Innovation Award. The views and conclusions contained in this document are those of the authors and should not be interpreted as representing the NSF, Sony, or the U.S. Government.

\cleardoublepage
\balance

{
    \small
    \bibliographystyle{ieeenat_fullname}
    \bibliography{IEEEabrv, main}
}

\end{document}